\documentclass{ecai}
\usepackage{latexsym}
\usepackage{amsfonts}
\usepackage{amsmath}
\usepackage{bm}

\usepackage{algorithm}%
\usepackage{algorithmicx}%
\usepackage{algpseudocode}%

\usepackage{graphicx}
\usepackage[caption=false]{subfig}
\usepackage{xcolor}
\usepackage{booktabs}       % professional-quality tables
\usepackage{url}            % simple URL typesetting
\usepackage{multirow}

\newcommand{\xvar}{\mathbf{x}}

\begin{document}

\begin{frontmatter}

\title{Revisiting the Robustness of the Minimum Error Entropy Criterion: A Transfer Learning Case Study}
% \author{\fnms{Anonymous}~\snm{Authors}}
\author[A]{\fnms{Luis Pedro}~\snm{Silvestrin}\orcid{0000-0002-5759-1986}\thanks{Corresponding Author. Email: l.p.silvestrin@vu.nl.}}
\author[A]{\fnms{Shujian}~\snm{Yu}\orcid{0000-0002-6385-1705}\thanks{Email: s.yu3@vu.nl}}
\author[A]{\fnms{Mark}~\snm{Hoogendoorn}\orcid{0000-0003-3356-3574}} 
% use of \orcid{} is optional

\address[A]{Vrije Universiteit Amsterdam}
% \address[B]{Short Affiliation of Second Author and Third Author}

\begin{abstract}
Coping with distributional shifts is an important part of transfer learning methods in order to perform well in real-life tasks.
However, most of the existing approaches in this area either focus on an ideal scenario in which the data does not contain noises or employ a complicated training paradigm or model design to deal with distributional shifts.
In this paper, we revisit the robustness of the minimum error entropy (MEE) criterion, a widely used objective in statistical signal processing to deal with non-Gaussian noises, and investigate its feasibility and usefulness in real-life transfer learning regression tasks, where distributional shifts are common. 
Specifically, we put forward a new theoretical result showing the robustness of MEE against covariate shift. We also show that by simply replacing the mean squared error (MSE) loss with the MEE on basic transfer learning algorithms such as fine-tuning and linear probing, we can achieve competitive performance with respect to state-of-the-art transfer learning algorithms. We justify our arguments on both synthetic data and 5 real-world time-series data.
\end{abstract}

\end{frontmatter}

\section{Introduction}
Robustness is an essential quality for machine learning models to cope with the challenges of real-world scenarios. 
Typical challenges where robustness is desired include the distributional shift between training and test data \cite{quionero-candela_2009}, noisy data \cite{song_TNNLS22}, and adversarial attacks \cite{akhtar_access2018}.
Distributional shifts can result in poor generalization performance, as the model may rely its decision on spurious correlations in the training set~\cite{ahuja2021invariance}, while noisy data, such as label noise or response variable noise, can further bias the resulting model.

Most of the robustness research in transfer learning focuses on covariate shift, a special case of general distributional shift in which only the distribution of input ($p(\mathbf{x})$) changes and the conditional distribution ($p(y|\mathbf{x})$) remains the same. 
These methods aim to learn models that are less sensitive to changes in the data distribution and can adapt to new environments. 
However, these approaches either focus on an ideal scenario in which the source and target domains are noise free, or are very complicated to implement, requiring extensive training or hyperparameter tuning.
For example, there are approaches based on adversarial training \cite{DANN_ganin_jmlr2016,mathelin_2021_IEEE_ICTAI} which is known for being difficult to converge, and approaches based on boosting \cite{tradaboost_daietal_icml2007,pardoe_2010_ICML} which also require extensive hyperparameter tuning of both the base estimators and the approach itself.

Research addressing the challenges of noisy data predominantly focuses on the classification setting, also known as label noise~\cite{song_TNNLS22}. 
However, there is notably less emphasis on regression problems.
For example, most of existing machine learning approaches simply use the mean-squared error (MSE) loss or the mean-absolute error (MAE) loss. The former implicitly assumes the noises (in the response variable) follow a Gaussian distribution, whereas the latter takes a Laplacian distributional assumption.  
Training a model using MSE or MAE is likely to negatively impact its performance on real-world data, especially when noise is non-Gaussian or non-Laplacian. Therefore, it is of paramount importance to design a transfer learning model such that it can handle a wide range of noise distributions in a non-parametric way (without distributional assumption on noises).

In this paper, we present an approach focusing on covariate shift in a realistic transfer learning regression scenario, where non-Gaussian noise is present.
We do so by combining the minimum error entropy (MEE) loss~\cite{erdogmus_TSP2002} - a widely used learning objective in statistical signal processing to deal with non-Gaussian noises - with classic deep transfer learning methods such as fine-tuning and linear probing. MEE has received lots of attention in signal processing and information theory literature, whereas its practical usage in machine learning, especially deep neural networks, is scarcely investigated due to the difficulty of entropy estimation~\cite{principe2010information}. Our work put forward a new theoretical result on the robustness of MEE, showing that it also encourages the robustness to covariate shift. 

We conduct comprehensive experiments on both synthetic data and $5$ real-world time-series data, showing that the simple combination strategy outperforms existing deep neural network based transfer learning approaches with complicated model design or training paradigms.
Time-series data serves as an ideal testing ground for our approach since it is commonly affected by covariate shift in many forms such as seasonal variation or sensor drift and contains measurement noises of unknown distributions.
The main contributions of this paper are summarized in the following points:

\begin{itemize}
    \item We provide a theoretical result showing that, besides its resilience to non-Gaussian noise in the response variable, MEE can also cope with covariate shift.
    \item We use the transfer learning regression setting to show empirically that by simply replacing the training loss with MEE the resulting model becomes more robust to both covariate shifts and response variable noise.
    \item We compare our approach with other state-of-the-art robust learning methods, and our method consistently outperforms them in multiple real-life time-series transfer learning tasks.
\end{itemize}

\section{Background and Related Work}
In this section, we introduce the prior knowledge about the MEE loss that we will base the rest of the paper on. 
We also cover the previous work on robust machine learning in general.

\subsection{Minimum Error Entropy Criterion}
We assume that the explanatory variable $\xvar$ takes values in a compact domain $\mathcal{X} \in \mathbb{R}^d$, the response variable $y$ takes values in the output space $\mathcal{Y} \in \mathbb{R}$, and 
\begin{equation}
    y = g^*(\xvar) + \epsilon
\end{equation}
where $g^*$ is the ground-truth target function and $\epsilon$ is the noise in the regression model.

The purpose of regression is to estimate $g^*(\xvar)$ according to a data set $\mathcal{D}=\{\xvar_i, y_i\}^N_{i=1}$ drawn independently from an unknown joint distribution $p(\xvar, y)$. 
Usually, a loss function $\mathcal{L}(g, (\xvar,y))$ is used to measure the performance of a hypothesis function $g:\mathcal{X}\rightarrow \mathcal{Y}$. 
For regression, the most used loss function is the mean-squared loss:
\begin{equation}
    \mathcal{L}_{\text{MSE}} = \frac{1}{N}\sum^N_{i=1}\left(y_i-g(\xvar_i)\right)^2 = \frac{1}{N}\sum^N_{i=1}e^2_i
\end{equation}
where $e_i = y_i - g(\xvar_i)$ is the prediction error for sample $(\xvar_i, y_i)$. 
The MSE minimizes the variance of the prediction error. 
Its optimality heavily depends on the Gaussianity of the data due to the use of a second-order statistic. Hence, the MSE solution may deviate significantly from the ground truth, especially in the presence of noise. 

Alternatively, one can obtain $g$ by minimizing the entropy of the prediction error $H(e)$, which is also called the minimum error entropy criterion~\cite{erdogmus_TSP2002}. 
If we instantiate $H(e)$ by R\'{e}nyi’s $\alpha$-order ($\alpha > 0$ and $\alpha \neq 1$) entropy functional \cite{renyi_1960}, the resulting objective of MEE becomes:
\begin{equation}
   \min H_\alpha(e) = \min \frac{1}{1-\alpha}\log \int p^\alpha(e)de,
\end{equation}
in which $p(e)$ is the probability distribution function (PDF) of prediction error $e$. Entropy is a functional of the PDF and measures the average information contained in that distribution. 
The basic idea of MEE is to reduce the uncertainty (entropy) of the discrepancies between models and data generating systems and improve the models’ predicting capability for unseen data \cite{erdogmus_TSP2002}. 
Compared to MSE, the R\'{e}nyi’s entropy takes into consideration all higher moments. 
Hence, the MEE can deal with outliers, heavy-tailed noise, or skewed noise distributions.

Practically, $\alpha = 2$ (\emph{a.k.a.}, quadratic R\'{e}nyi entropy) is the most popular choice, as it can be elegantly estimated by the kernel density estimator (KDE) \cite{parzen_62}. 
In this case, we have:
\begin{equation}
    \min H_2(e) \iff \max \int p^2(e) de,
\end{equation}
and
\begin{equation}
    \hat{p}(e) = \frac{1}{N} \sum^N_{i=1} \kappa_\sigma(e-e_i),
\end{equation}
where $\kappa_\sigma$ is a Gaussian kernel function with width $\sigma$.

Hence, the empirical MEE loss can be expressed as:
\begin{equation} \label{eq:empirical_mee}
    \mathcal{L}_{\text{MEE}} = \max \int \hat{p}^2(e) de = \max \frac{1}{N^2}\sum^N_{i=1} \sum^N_{j=1} \kappa_{\sqrt{2}\sigma} (e_i-e_j).
\end{equation}

Although there is a series of works on the theory and applications of MEE, such as \cite{hu_JMLR13,chen_TNNLS18,guo2020distributed}, just to name a few, they only investigate or utilize the robustness of MEE against non-Gaussian noises in $y$. 
Distinct from these works, one of the motivations of our paper is to point out and systematically investigate the robustness of MEE against distributional shift (i.e., $p(\mathbf{x},y)$ differs in training and test set), which, to the best of our knowledge, has not been done yet. Moreover, instead of estimating $H_\alpha(e)$ by KDE as shown in Eq.~(\ref{eq:empirical_mee}), we suggest the use of the matrix-based R\'enyi's $\alpha$-order entropy functional~\cite{shujian_AAAI21, sanchez_rao_principe_IEEE2014} to measure $H_\alpha(e)$, which avoids density estimation and thus more suitable for complex data and deep neural networks~\cite{yu2019understanding}.

\subsection{Robust Machine Learning with Covariate Shift}
Literature on robust learning under distributional shift mainly focused on dealing with covariate shift, and assumes the training and test data is noise free.
Here we list the most recent and relevant works on covariate shift split as those assuming the existence of a target dataset (transfer learning and domain adaptation), and those without that assumption.

\textbf{Transfer Learning and Domain Adaptation:}
The majority of transfer learning approaches proposed in the literature categorize according to the differences in the input domains (homogeneous or heterogeneous) or according to the methodology used to bridge the gap between source and target domains \cite{Weiss2016_survey}.
Homogeneous transfer learning assumes that both source and target inputs come from the same feature space and therefore have the same dimensionality, whereas for heterogeneous transfer learning they come from distinct spaces.
In this work, we are concerned with the homogeneous category.
Within the homogeneous transfer learning literature, the most prominent methodologies fall into two categories: feature-based and instance-based.
Feature-based approaches try to mitigate the covariate shift problem by mapping the inputs of the source domain (or both source and target domains) to a feature space where their distribution matches with the target input domain distribution.
Instance-based approaches, on the other hand, propose to solve the same issue by weighting the source samples according to their relevance to the target task so, intuitively, source samples that are distant in the target distribution will become less important during training.

In the instance-based category, we have TrAdaBoostR2 (TRB) \cite{pardoe_2010_ICML}: a combination of TrAdaBoost \cite{tradaboost_daietal_icml2007}, a transfer learning variation of AdaBoost, and AdaBoostR2 \cite{adaboostr2_druker_icml97}, which is AdaBoost adapted for regression.
It adapts both algorithms to transfer learning regression by taking into account that the base learner errors are unbounded, differently from classification, and it bridges the covariate shift by down-weighting source samples that are far from the target distribution.
Another more recent instance-based approach is WANN \cite{mathelin_2021_IEEE_ICTAI}, a neural network trained by minimizing an adversarial loss derived from an upper bound of the target generalization error.
Its proposed training method includes an auxiliary network that predicts weights to maximize the error of the final model on the source samples, thus indicating which samples are less relevant for the target task.

Among the feature-based transfer learning approaches, there is an adversarial algorithm that learns a new feature representation to align source and target domains by minimizing the margin disparity discrepancy (MDD) proposed by the authors \cite{MDD_zhang_icml19}.
Domain-adversarial neural networks (DANN) \cite{DANN_ganin_jmlr2016} is a seminal method applying adversarial learning for better feature representations across domains.
It trains three neural networks simultaneously: one for feature mapping ($G_f$), one for classification ($G_y$), and the last one ($G_d$) is trained to predict whether the representation given by $G_f$ was produced from a source or target sample.
The adversarial aspect of it comes from the min-max game between $G_f$, which optimizes to fool $G_d$.
In the end, $G_f$ should learn a unified representation for both source and target samples which is indistinguishable to $G_d$ and easily separable for the classifier $G_y$.

All the aforementioned works propose a complex algorithmic solution to the covariate shift problem in transfer learning and can be difficult to optimize while also introducing extra hyperparameters.
They differ fundamentally from our approach, which focuses on replacing the MSE with MEE as the training objective.
Since we only change the loss function, our approach can be easily implemented on the existing fine-tuning and linear probing algorithms.
Nevertheless, we compare the original TRB and WANN algorithms with our approach in multiple time-series transfer learning regression tasks in our experiments.
We choose TRB and WANN over the rest since they showed better performance in several transfer learning regression benchmarks \cite{mathelin_2021_IEEE_ICTAI}.

\textbf{Robust learning without target data:} 
Another new and less investigated area of research proposes approaches for robustness to covariate shift without assuming any specific knowledge target domain (i.e., no labeled or labeled samples from test set). Note that, this setup is also different to domain generalization (e.g., \cite{ahuja2021invariance}), in which there are multiple related source domains during training. See also supplementary material for comparison with respect to domain generalization.
Anchor regression \cite{rothenhausler_jrssb21} is a least-squares-based method that includes exogenous variables in the regression model to account for possible causal interventions in the data.
Another work proposes \cite{subbaswamy_aistats19} an algorithm robust to changes in the data-generating process.
It requires the process to be modeled as a causal graph and the changes from source to target domains should be known so that their algorithm is able to take it into account.
Although these approaches ensure robustness to changes in the inputs, they are also limited to linear problems so it is less suitable for many real-life regression tasks.

A recent work \cite{greenfield_icml2020} shows that by training a model by minimizing the Hilbert Schmidt Independence Criterion (HSIC) the resulting model can be more robust to covariate shift.
The HSIC \cite{HSIC_Gretton_2005} is originally studied as an independence measure for random variables, but subsequent work \cite{Mooij_ICML2009} shows that it is also suitable as a loss function by using it to minimize the statistical dependence between covariates and labels.
This work motivates our approach to investigate the covariate shift robustness of the MEE criterion since, in the regression case, it is related to minimizing the mutual information between covariates and labels which, in turn, is also a statistical dependence measure.

\section{Transfer Learning with MEE}
In transfer learning, we assume a small target dataset $\{\xvar_{iT}, y_{iT}\}_{i=1}^{N_T}$, with inputs $\xvar_{iT}$ and labels $y_{iT}$ drawn from the target distribution $p_T(\xvar, y)$, and a large source dataset $\{\xvar_{iS}, y_{iS}\}_{i=1}^{N_S}$ ($N_S \gg N_T$) with inputs $\xvar_S$ and labels $y_S$ drawn from the source distribution $p_S(\xvar, y)$. 
The final goal is to build a model that generalizes to new unseen samples from $p_T$.
Since the target data is limited, transfer learning proposes to use it combined with the extra source data to create such model.
A common challenge practitioners face, besides the lack of target data, is bridging distribution shifts between $p_T$ and $p_S$.
There are three types of shifts that can occur in practice: covariate shift ($p_T(\xvar) \neq p_S(\xvar)$), label distribution shift ($p_T(y) \neq p_S(y)$), and labeling function shift ($p_T(y|\xvar) \neq p_S(y|\xvar)$), and there is a plethora of approaches for each of them \cite{pan_survey_tl_ieee2010}.

In this section, we first provide a new perspective on the robustness of MEE to covariate shift, which motivates our study.
We then describe our main algorithm that integrates MEE into two basic transfer learning paradigms. 
Finally, we elaborate on the implementation details, including the way to estimate entropy, the way to compensate for the model bias, and the way to estimate the kernel size.

\subsection{Theoretical Justification on the Robustness of MEE to Covariate Shift}

The robustness of MEE against non-Gaussian noises has been extensively investigated in previous literature. We refer interested readers to \cite{chen_TNNLS18,chen_SP2010,chen_AAS2009,hu_JMLR13} for more thorough analysis. We also summarize in the supplementary material two key points in this context.

The robustness of MEE against covariate shift is easy to understand. Note that we have:
\begin{align}\label{eq:proof_mee}
    \min H(e)   & \iff \min H(e) - H(y|\xvar) \nonumber \\
                & = \min H(e) - H(e + f_\theta(\xvar)|\xvar) \nonumber \\
                & = \min H(e) - H(e|\xvar) \nonumber \\
                & = \min I(\xvar;e)
\end{align}
The first line is due to the fact that the conditional entropy $H(y|\xvar)$ is a constant value that only depends on the training data; the third line is by the property that given two random variables $\xi$ and $\eta$, then for any measurable function $h$, we have $H(\xi|\eta) = H(\xi + h(\eta)|\eta)$ \cite{mackay_info_theory_2002}.

Hence, minimizing the error entropy actually encourages the minimum dependence between $\xvar$ and $e$. In other words, the distribution of input variable $\xvar$ is independent of the distribution of prediction error $e$. Since the prediction performance is characterized by $p(e)$\footnote{A good predictor is expected to have a concentrated error distribution with zero mean.}, it also suggests that the predictor performance was not impacted by the change of $p(\xvar)$, i.e., covariate shift.
Therefore the MEE is an ideal loss function in a transfer learning scenario where $p_S(\xvar) \neq p_T(\xvar)$ and $p(y|\xvar)$ is the same in both source and target domains.

Note that, Greenfeld and Shalit~\cite{greenfield_icml2020} firstly observed and rigorously proved that the HSIC between $p(e)$ and $p(\xvar)$ is an upper bound of the worst-case loss in the target 
domain in case of covariate shift.
%the robustness against covariate shift by the HSIC loss (i.e., $\text{HSIC}(\mathbf{x};e)$). 
Our simple proof in Eq.~(\ref{eq:proof_mee}) generalizes the arguments in \cite{greenfield_icml2020}, showing that any independence measures can be used here (rather that just HSIC). Additionally, \cite{greenfield_icml2020} does not discuss the close relationships between $\min I(\xvar;e)$ and MEE; and does not systematically investigate the performance and utility of MEE in a practical transfer learning scenario.

\subsection{Integration of Minimum Error Entropy and Transfer Learning}

In this paper, we focus on two widely used transfer learning techniques: fine-tuning and linear probing.
Fine-tuning is a popular approach to transfer learning and has shown great success in several real-life tasks \cite{Oquab_2014_CVPR}.
It consists of readjusting all the weights of a neural network pre-trained with the source dataset through gradient descent by minimizing a given loss function $\mathcal{L}$ using the target data.
We assume that the pre-trained neural network architecture $g(\xvar; w, \theta) = w^\top f(\xvar;\theta)$ is split into the feature extracting layers $f(\xvar; \theta)$ with weights $\theta_S$ and the regression layer with weight matrix $w_S$.
Since the source dataset is usually larger and more diverse, the pre-trained model could already have a better performance in the target domain than a randomly initialized model, therefore it works as a "warm start" and makes training easier since the model will need fewer data and epochs to converge.
For this approach to work successfully, it requires a certain degree of similarity between the source and target tasks.
If both tasks are similar enough, it is likely that fine-tuning will result in a better model for the target task, otherwise negative transfer can happen \cite{Weiss2016_survey}, meaning that the generalization performance on the target task will be hurt.

In linear probing, we freeze the parameters $\theta_S$ of the feature extracting layers $f$ of the pre-trained source model and we update only the last layer's weights $w_S$.
This way, we are reusing the same features learned by the source model, so we only need to adapt the last layer to the target task.
The idea behind it is that we can tap into the neural network capabilities of learning to extract general meaningful features from large amounts of data. 
If the source dataset is large and contains a wide variety of samples, then the features extracted by the source model are likely to be relevant also for the target task.
On top of that, training only the last layer can be preferable over fine-tuning all the layers since the latter option can distort the features learned from the source dataset \cite{kumar_iclr23}.

Fine-tuning and linear probing are generally preferred over other more complex approaches since they can be easily combined with any neural network architecture, data types, and loss functions, while also being able to give good results \cite{liu_Comput_survey_NLP,Hu_2022_CVPR}.
They are commonly used for neural network transfer learning combined with the cross-entropy loss for classification or the MSE for regression.
We propose to improve the regular fine-tuning and linear probing algorithms for regression tasks by combining it with the MEE loss.
We do that by instantiating the MEE loss in the place of the MSE and by including two extra steps required by the MEE: the kernel size estimation and the model bias correction.

Algorithms \ref{alg:fine-tuning} and \ref{alg:linear-probing} describe in pseudo-code respectively our novel approaches based on fine-tuning and linear probing.
The main suggested modifications are annotated with comments.
Before the training starts, we compute the RBF kernel width $\sigma$ as the median of the pair-wise distance between the residuals of the source model on the target training samples.
Once the network parameters are optimized using MEE, we compute the model bias $b$ as the average of its residuals on the training target data.
In the following sections we go into detail of how we compute the MEE loss, as well as the justification for the extra steps that it requires.

\begin{algorithm}[tb]
   \caption{Fine-tuning with MEE}
   \label{alg:fine-tuning}
\begin{algorithmic}[1]
   \State {\bfseries Input:} target dataset $(\xvar_T, y_T)$, source feature extractor parameters $\theta_S$, source regressor parameters $w_S$, learning rate $\eta$, number of epochs $M$.
   \State $\theta_0 \leftarrow \theta_S$ 
   \State $w_0 \leftarrow w_S$
   \State $e_i \leftarrow y_{iT}- w_S^\top f(\xvar_{iT}; \theta_S), \forall i \in \{1, ... N\}$ \hfill\Comment{compute residuals}
   \State $\sigma \leftarrow \text{median}(\{(e_i - e_j)^2\}^N_{i,j=1})$    \hfill\Comment{kernel size estimation}
    \For{$i \leftarrow 1$ to $M$}
        \State $\theta_i \leftarrow \theta_{i-1} + \eta \nabla_{\theta}  \mathcal{L}_{\text{MEE}}(w_{i-1}, \theta_{i-1})$   \hfill\Comment{MEE}
        \State $w_i \leftarrow w_{i-1} + \eta \nabla_{w} \mathcal{L}_{\text{MEE}}(w_{i-1}, \theta_{i-1})$   \hfill\Comment{MEE}
    \EndFor
    \State $b \leftarrow  \frac{1}{N}\sum^N_{i=1}(y_i - w_M^\top f(\xvar_T; \theta_M))$ \hfill\Comment{bias correction}
   \State {\bfseries Output:} fine-tuned parameters $\theta_M, w_M$ and bias $b$
\end{algorithmic}
\end{algorithm}

\begin{algorithm}[tb]
   \caption{Linear probing with MEE}
   \label{alg:linear-probing}
\begin{algorithmic}[1]
   \State {\bfseries Input:} target dataset $(\xvar_T, y_T)$, source feature extractor parameters $\theta_S$, source regressor parameters $w_S$, learning rate $\eta$, number of epochs $M$.
   % \State $\theta_0 \leftarrow \theta_S$ 
   \State $w_0 \leftarrow w_S$
   \State $e_i \leftarrow y_{iT}- w_S^\top f(\xvar_{iT}; \theta_S), \forall i \in \{1, ... N\}$ \hfill\Comment{compute residuals}
   \State $\sigma \leftarrow \text{median}(\{(e_i - e_j)^2\}^N_{i,j=1})$                    \hfill\Comment{kernel size estimation}
   \For{$i \leftarrow 1$ to $M$}
       % \State $\theta_i \leftarrow \theta_{i-1} + \alpha \nabla_{\theta} \mathcal{L}(w_{i-1}^\top f(\xvar_T; \theta_{i-1}), y_T)$
       \State $w_i \leftarrow w_{i-1} + \eta \nabla_{w} \mathcal{L}_{\text{MEE}}(w_{i-1}, \theta_S)$ \hfill\Comment{MEE}
   \EndFor
    \State $b \leftarrow  \frac{1}{N}\sum^N_{i=1}(y_i - w_M^\top f(\xvar_T; \theta_M))$ \hfill\Comment{bias correction}
   \State {\bfseries Output:} fine-tuned parameters $w_M$ and bias $b$
\end{algorithmic}
\end{algorithm}

\subsubsection{Matrix-based Implementation of Minimum Error Entropy}
As mentioned in the previous section, the MEE loss has interesting robustness properties, but its KDE version described in Eq.~(\ref{eq:empirical_mee}) is only suitable for low-dimensional problems, and it is difficult to select the appropriate kernel function.
For that reason, in this paper we implement MEE using the matrix-based version \cite{sanchez_rao_principe_IEEE2014} of the quadratic R\'{e}nyi's entropy of the model residuals $e$:
\begin{equation} \label{eq:matrix-based-MEE}
    \mathcal{L}_{\text{MEE}}(w, \theta) = \frac{1}{2} \log_2 \left[ \sum^N_{i=1} \lambda_i(A)^2\right]
\end{equation}
where $A$ is a normalized positive definite matrix computed as $A_{ij} = \frac{1}{N}\frac{K_{ij}}{\sqrt{K_{ii}K_{jj}}}$ and $\lambda_i(A)$ is the i-th eigenvalue of A.
$K$ is the Gram matrix obtained by evaluating a positive-definite kernel $\kappa$ on the model's residuals $e_i = y_i - g(\xvar_i)$ on each pair of training data point.
The kernel used is the radial basis function (RBF), so $\kappa$ and $K$ are expressed as:
\begin{align*}
    K_{ij} &= \kappa_{\sigma} (e_i, e_j), \\
    \kappa_\sigma(\xvar, y) &= \exp \left(-\frac{\Vert \xvar-y \Vert^2}{2\sigma^2}\right).
\end{align*}
By instantiating the model $g(\xvar)$ with the feature extractor $f$ plus a linear output layer with parameter $w$ and the target dataset $(\xvar_T, y_T)$, the Gram matrix $K$ becomes:
\begin{equation*}
    K_{ij} = \kappa_\sigma(y_{iT}-w^\top f(\xvar_{iT};\theta), y_{jT} - w^\top f(\xvar_{jT}; \theta))
\end{equation*}

This matrix-based implementation is ideal since it is differentiable and can be computed in tractable time.
In this paper, for the first time, we employ it to transfer learning tasks by coupling it to the fine-tuning and linear probing frameworks. 
Next, we explain in detail and justify the model bias correction and the computation of the kernel size which are important steps in our algorithms.

\begin{figure*}
    \centering
    \subfloat[Shifted exponential noise.]{
    \includegraphics[width=0.3\linewidth]{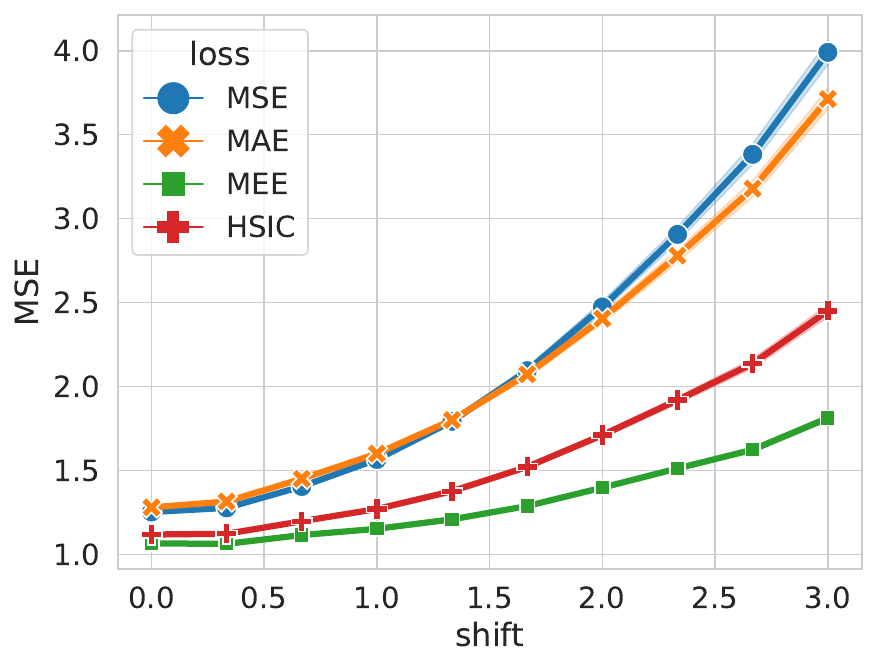}
    \label{fig:covshift_exponential_noise}
    }\qquad
    \subfloat[Mixed Gaussian noise.]{
    \includegraphics[width=0.3\linewidth]{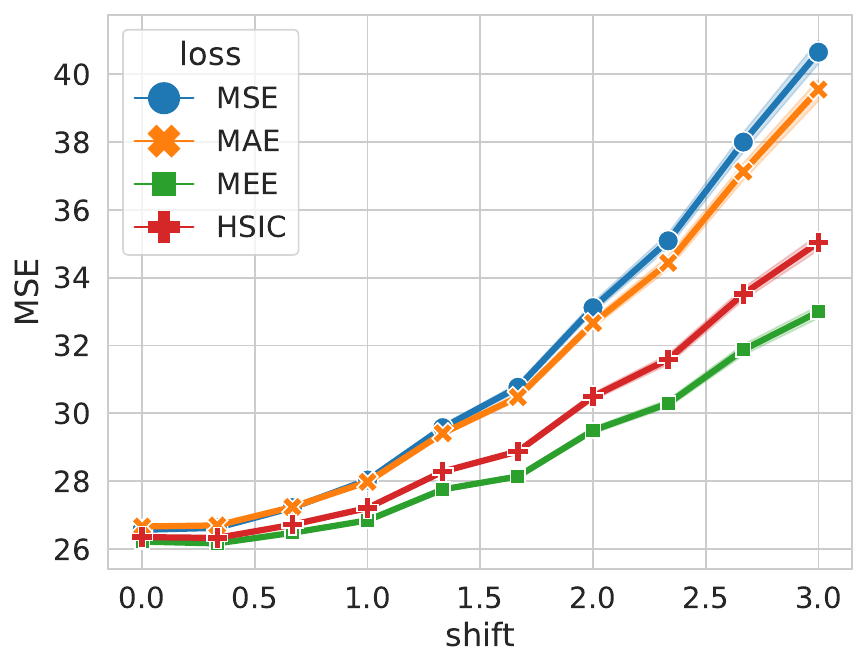}
    \label{fig:covshift_gaus_noise}
    }\qquad
    \subfloat[Laplace noise.]{
    \includegraphics[width=0.3\linewidth]{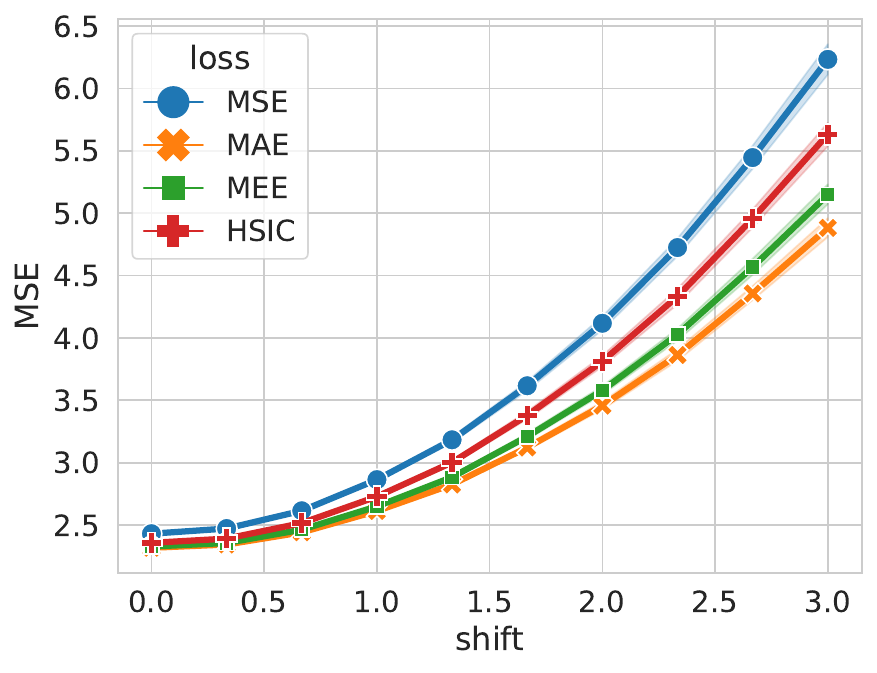}
    \label{fig:covshift_laplace_noise}
    }
    \caption{Comparison of the squared error of regression models trained using different loss functions and evaluated on data with increasing covariate shift for different types of response variable noise. The bands indicate the standard deviation.}
    \label{fig:covshift}
\end{figure*}

\subsubsection{Correcting the Model Bias}
\label{sec:model_bias}
An important difference between MEE and other loss functions such as MSE is that it give the same loss value for models with different errors.
More specifically, the error entropy $H(g(\xvar) - y)$ is the same for any models $g_1(\xvar)$ and $g_2(\xvar)$ that differ only by a constant $C$.
This can be easily seen by defining the errors of each model respectively as $\epsilon_1$ and $\epsilon_2 = \epsilon_1 + C$.
Therefore the error entropy of each model will then only differ by the integral of the error PDFs $p(\epsilon_1)$ and $p(\epsilon_2)$, which in fact are the same.
This can be an issue because, even if the model's error entropy has converged to a minimum after training, its predictions might still deviate from the ground truth by a bias constant.

To solve this problem, we estimate the model's bias $b$ as:
\begin{equation*}
    b = \frac{1}{n}\sum^n_{i=1}(y_i - g(\xvar_i))
\end{equation*}
The bias $b$ is then added to the output to correct the model's predictions, so the corrected model becomes $g_{\text{corrected}}(\xvar) = g(\xvar) + b$.

\subsubsection{Computing the Kernel Size}
\label{sec:kernel_size}
The matrix-based implementation of MEE used in this paper relies on the RBF kernel to approximate the PDF of the error in a tractable way.
It also introduces the kernel size (or width) $\sigma$ as an extra hyperparameter to tune in order to obtain an accurate entropy estimation.
It is important to have a proper value for $\sigma$ since otherwise, it can lead to convergence problems during training. 
A too large $\sigma$ will result in an all-ones Gram matrix $K$, thus its eigenvalues in Eq. (\ref{eq:matrix-based-MEE}) will tend to zero, while if $\sigma$ is too small then $K$ will approach the identity and the eigenvalues will approach $1/N$.
Furthermore, if $\sigma$ is small enough (e.g., $\sigma = 0$) the entropy of error will tend to be a fixed value ($\log_2 N$) and is maximized, which also violates the goal of minimizing the error entropy.
In either case, the landscape of Eq. (\ref{eq:matrix-based-MEE}) over the parameter space will flatten and the training is likely to fail.
In our empirical experiments, we calculate the $\sigma$ to be the median of the pair-wise euclidean distance between the model errors prior to training:
\begin{equation}
    \sigma = \text{median}(\{(e_i - e_j)^2\}^N_{i,j=1})
\end{equation}
This heuristic is also called the median-rule and has been used in practice by previous kernel learning literature~\cite{gretton2012kernel,yu_TPAMI2020,jenssen2009kernel} and also results in good training convergence in our experiments.

\section{Experiments}
\label{sec:experiments}
To empirically evaluate our results about the robustness to covariate shift of the MEE as a loss function, we conduct experiments with both real and synthetic regression datasets, the former focusing on time-series data as argued before.
In the first part, we use synthetic regression data to verify empirically our theoretical hypothesis about the robustness of the MEE loss to non-Gaussian noises and covariate shift.
In the remaining parts, we use real-life time-series transfer learning regression datasets, which, as argued before, are highly suitable for studying distributional shift robustness of transfer learning methods. 
We use them to compare the performance of our approach with other state-of-the-art transfer learning algorithms. \footnote{Our code is available at \url{https://github.com/lpsilvestrin/mee-finetune}.}

\subsection{Synthetic Linear Regression Experiment}
In this experiment, we evaluate the robustness of the MEE loss to covariate shift using synthetic data so we can fully control the amount of covariate shift between source and target data.
We compare MEE with the popular MSE and MAE, as well as the state-of-the-art HSIC, gaining detailed insights about the performance of each loss function.
We use a linear model to generate the data to ensure that the gradient descent optimization will converge to a global optimum.

We generate the data of both source and target domains using a linear model $y=\theta^\top \xvar + \epsilon$, where $y$ is the response variable, $\xvar$ are the inputs, $\theta$ are the regression coefficients and $\epsilon$ is the additive noise.
We randomly sample the coefficients $\theta$ once from $\mathcal{N}(0, 0.1)$ and keep them always fixed. 
The coefficients $\theta$ and the distribution of $\epsilon$ remain the same for both source and target datasets.
We emulate different degrees of covariate shift by simulating a source dataset with fixed input distribution $p_S(\xvar)$, then simulating multiple target datasets with increasing distribution shift in $p_T(\xvar)$.
The source inputs $\xvar_S$ are sampled from a uniform distribution in the real interval $[-1,1]^{100}$, while the target inputs $\xvar_T$ are sampled from a normal distribution $\mathcal{N}(\mu_T, 1)$, and we vary $\mu_T$ from 0 up to 3.
The covariate shift increases as the mean of $\xvar_T$ changes: when $\mu_T=0$, $p_S(\xvar)$ and $p_T(\xvar)$ have the most overlap, and as we increase $\mu_T$, $p_T(\xvar)$ will sample more and more inputs that are outside the support of $p_S(\xvar)$.
In order to also compare the robustness of the losses to different noise distributions, we repeat the experiment using shifted exponential noise, mixed Gaussian noise, and Laplace noise.
Further implementation details are discussed in the supplementary material.

The results in Figure \ref{fig:covshift_exponential_noise} and Figure \ref{fig:covshift_gaus_noise} show the mean-squared error of the regression models trained with different loss functions in the y-axis and the distance between the means of the input distributions of the source and target datasets in the x-axis.
As the target input distribution shifts, MEE's error showed a significantly slower increase compared to other approaches.
In the shifted exponential noise case (Figure \ref{fig:covshift_exponential_noise}), where the difference is striking even when compared to the state-of-the-art HSIC.
MEE is only outperformed by MAE in the Laplace noise case (Figure \ref{fig:covshift_laplace_noise}), which is expected since the model trained with MAE is the maximum likelihood solution to this case. 
In all cases, MEE already has lower error than MSE with minimal shift, showing that it can handle the change from uniformly to normally distributed inputs better.
Overall, this experiment confirms our theoretical claims about the robustness to covariate shift and, additionally, shows its resilience to different types of response variable noise.

\begin{figure}
    \centering
    \includegraphics[width=0.9\linewidth]{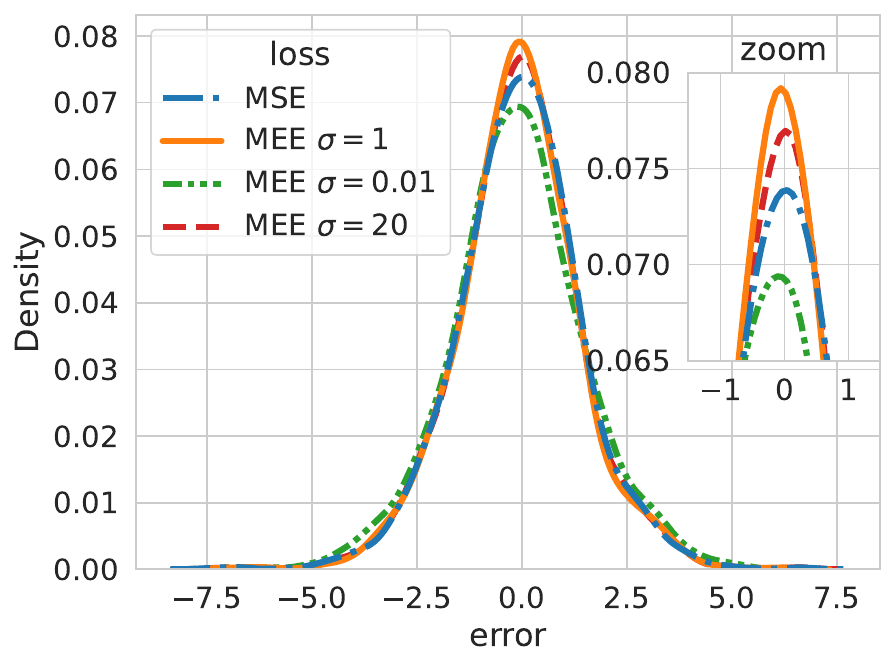}
    \caption{Comparison of error densities using MEE with different kernel sizes.}
    \label{fig:kernel_size}
\end{figure}

In Figure~\ref{fig:kernel_size} we use the dataset simulated with Laplace noise to visualize the effect of the kernel size $\sigma$ on the spread of the error distribution.
We can see that by picking it as the median ($\sigma=1$) the errors are more concentrated around zero\footnote{The error entropy is minimized if the probability of one state dominates, thus forming a highly concentrated distribution.}, while the larger or smaller choices result in a more spread distribution.
Additionally, when reducing the size causes substantial degradation of the model such that its performance is even inferior to that of the MSE.
This confirms that the procedure explained in Section \ref{sec:kernel_size} eliminates the need for practitioners to manually tune $\sigma$.

\subsection{Time-series Transfer Learning}
In this experiment, we verify the performance of our MEE version of fine-tuning and linear probing using 5 real-life time-series regression transfer learning tasks using deep neural networks.
We conduct an ablation study comparing our algorithm with the versions using the MSE, MAE, and HSIC.
We also compare our method with TRB and WANN which are state-of-the-art transfer learning regression algorithms.

\subsubsection{Datasets}

We compare all the training losses and transfer learning approaches using 5 real-life time-series regression tasks based on 3 datasets from the popular Monash \cite{godahewa2021monash} and UCI \cite{uci_dataset_repository} repositories: the Nasa Turbofan, the Beijing air quality and the bike sharing datasets.
The Nasa Turbofan data contains 4 datasets where engines operate under different conditions, providing distributional shifts between each other.
We select one of them as a source dataset (NTS), and the remaining as target datasets (NT1, NT2, and NT3).
For the Beijing air quality dataset, we use the first year of measurements (2013) as source data (PMS) and the early months of the last year (2017) as target training data (PMT).
The models are tested in the remaining data from 2017.
For the bike sharing dataset, we separate data from fall, winter, and spring as the source dataset (BKS) while the target data is from summer, and the training target dataset contains information only from the early days of summer.

A summary of all datasets is listed in Table \ref{tab:dataset_stats}.
Further details about the distributional shifts on each dataset are discussed in the supplementary material.

\begin{table}
\caption{The size of the train and test sets, the length of the time window and the amount of features of each dataset used in our experiments.}
\label{tab:dataset_stats}
\begin{center}
\begin{tabular}{lllcl}
\hline
\rule{0pt}{12pt}
    \textbf{Dataset} & $n_{\text{train}}$ & $n_{\text{test}}$  & window size & features 
\\
\hline
\\[-6pt]
NTS & 14,432    & 3,299   & 30 & 14  \\
NT1 & 1,678     & 44,541  & 30 & 14  \\
NT2 & 2,225     & 19,595  & 30 & 14 \\
NT3 & 2,261     & 51,767  & 30 & 14 \\
PMS & 4,000     & 1,000   & 24 & 9   \\
PMT & 200       & 1,000   & 24 & 9   \\
BKS & 5,143     & 1,266   & 24 & 10  \\
BKT & 403       & 1,681   & 24 & 10  \\
\hline
\\[-6pt]
\end{tabular}
\end{center}
\end{table}

\subsubsection{Linear Probing and Fine-tuning Experiments}
There are two steps where MEE can be applied when using fine-tuning (or linear probing) for transfer learning: the pre-training phase and the fine-tuning (or linear probing) phase itself.
The classic setup for regression is to use the MSE in both steps, therefore it is also selected as a baseline.
We conduct an ablation study with three experiments where we vary the loss functions in the fine-tuning and linear probing phases. 
In two of them, we keep the pre-training loss fixed as the MSE and in one experiment we vary the loss on both the pre-training and fine-tuning phases:
\begin{itemize}
    \item \textbf{Fix-vary with linear probing} We fix the pre-training loss and we vary linear probing loss.
    \item \textbf{Fix-vary with fine-tuning:} We fix the pre-training loss and vary the fine-tuning loss.
    \item \textbf{Vary-vary with fine-tuning:} We vary both the pre-training and fine-tuning loss.
\end{itemize}
In the linear probing setup, we evaluate the capacity of each loss to use the features learned by the pre-trained source model for predicting the target labels.
The fine-tuning-only setup compares how each loss can adapt the source model to each target task.
Finally, the pre-training plus fine-tuning setup compares the loss functions in a complete transfer learning cycle, from building a general source model to tuning it down to the target prediction problem.

For the source models, we use a temporal convolutional neural network (TCN) as the architecture for modeling sequential data.
This type of neural network is known for outperforming other classical time-series architectures such as the LSTM in many benchmark datasets \cite{bai_TCN_eval_2018}.
Further details of the hyperparameter choices are discussed in the supplementary material.

The HSIC is implemented with RBF kernels, as reported by the original paper \cite{greenfield_icml2020}. 
We select the kernel size the same way we do for MEE (Section \ref{sec:kernel_size}): we compute the median of the pairwise distance matrices for inputs and labels for each training dataset.
Table \ref{tab:rbf_kernel_size} contains the kernel sizes selected for covariates and for response variables for each dataset.
We repeat all the runs with all models and loss functions 20 times with different weight initializations and different training and validation samples, and we compare their results on the target test datasets.
The significance of the results is validated through a paired Wilcoxon test with an adjusted p-value of 0.05.

\begin{table}
\caption{RBF kernel size per dataset for both the covariates ($\sigma_X$) and the response variable ($\sigma_Y$).}
\label{tab:rbf_kernel_size}
\begin{center}
\begin{tabular}{lllllllll}
\hline
\rule{0pt}{12pt}
   &     NT1 & NT2 & NT3 & PMT & BKT &  NS & PMS & BKS
\\
\hline
\\[-6pt]
 $\sigma_Y$ &  0.5 & 1 &  0.5 &  0.3  &  0.8 &  0.3 & 0.5 &  0.3  \\
 $\sigma_X$ & 800  & 450 &  800 &  300   &  200 & 400  & 300 &  250  \\ 

\hline
\\[-6pt]
\end{tabular}
\end{center}
\end{table}

In the comparison using linear probing of a source model trained with the MSE (Table \ref{tab:linear_probing}), MEE displays significantly lower error than all other approaches in 4 out of 5 datasets.
Only for the NT2 dataset, there is no significant difference between the performances of MSE, MAE, and MEE.
This result is promising because it shows that by using MEE we are getting more from the features learned from the source dataset.

\begin{table}
\caption{\textbf{Fix-vary with linear probing:} average target squared error obtained from using different loss functions.}
\label{tab:linear_probing}
\begin{center}
\begin{tabular}{lllll}
\hline
\rule{0pt}{12pt}
    \textbf{Data} &  MSE-MSE & MSE-MAE & MSE-HSIC & MSE-MEE \\

\hline
\\[-6pt]
BKT & 0.25 $\pm$ 0.029 & 0.26 $\pm$ 0.04 & 0.26 $\pm$ 0.031 & \underline{0.24 $\pm$ 0.031} \\
NT1 & 0.87 $\pm$ 0.016 & 0.9 $\pm$ 0.0092 & 0.89 $\pm$ 0.025 & \underline{0.87 $\pm$ 0.018} \\
NT2 & \underline{0.48 $\pm$ 0.024} & \underline{0.48 $\pm$ 0.029} & 0.54 $\pm$ 0.047 & \underline{0.49 $\pm$ 0.025} \\
NT3 & 0.72 $\pm$ 0.022 & 0.73 $\pm$ 0.018 & 0.73 $\pm$ 0.015 & \underline{0.7 $\pm$ 0.0092} \\
PMT & 0.57 $\pm$ 0.039 & 0.56 $\pm$ 0.036 & 0.63 $\pm$ 0.038 & \underline{0.54 $\pm$ 0.034} \\
\hline
\\[-6pt]
\end{tabular}
\end{center}
\end{table}

Regarding the experiments using MEE only in the fine-tuning phase (Table \ref{tab:finetune-results}), we see that it outperforms all other losses in two datasets, and is on par with the best-performing (MSE) on two other datasets.
It is only outperformed in one dataset, where MAE and HSIC have lower error, but overall it confirms the be a more robust loss than MSE.
When using MEE for both pre-training and fine-tuning (Table \ref{tab:finetune+pretrain}), for two datasets it has lower error than the other losses.
Only for PMT and NT2, it results in higher errors compared to MSE and MAE, respectively.
However, MEE always has significantly better performance in more transfer learning tasks than each other loss in the individual comparison.
Summing up, the results confirm that our approach has more resilience to arbitrary distribution shifts encountered in real-life datasets compared to the counterparts using other popular training losses such as MAE and MSE, and even state-of-the-art losses such as HSIC.

\begin{table}
\caption{\textbf{Fix-vary with fine-tuning:} average target squared error obtained from using different loss functions. The pre-training loss function is always the MSE. For BKT and PMT, there was no significant difference between MEE and each other loss function.}
\label{tab:finetune-results}
\begin{center}
\begin{tabular}{lllll}
\hline
\rule{0pt}{12pt}
    \textbf{Data} & MSE-MSE & MSE-MAE & MSE-HSIC & MSE-MEE \\    
\hline
\\[-6pt]
BKT & 0.16 $\pm$ 0.015 & 0.17 $\pm$ 0.018 & 0.16 $\pm$ 0.018 & 0.16 $\pm$ 0.012 \\
NT1 & 0.47 $\pm$ 0.008 & 0.48 $\pm$ 0.008 & 0.48 $\pm$ 0.008 & \underline{0.46 $\pm$ 0.007} \\
NT2 & 0.59 $\pm$ 0.013 & \underline{0.58 $\pm$ 0.012} & \underline{0.58 $\pm$ 0.012} & 0.59 $\pm$ 0.009 \\
NT3 & 0.45 $\pm$ 0.007 & 0.45 $\pm$ 0.007 & 0.45 $\pm$ 0.007 & \underline{0.44 $\pm$ 0.006} \\
PMT & 0.44 $\pm$ 0.042 & 0.44 $\pm$ 0.038 & 0.47 $\pm$ 0.034 & 0.46 $\pm$ 0.024 \\
\hline
\\[-6pt]
\end{tabular}
\end{center}
\end{table}

\begin{table}
\caption{\textbf{Vary-vary with fine-tuning:} average target squared error obtained from using different loss functions.}
\label{tab:finetune+pretrain}
\begin{center}
\begin{tabular}{llllll}
\hline
\rule{0pt}{12pt}
    \textbf{Data} & MSE-MSE & MAE-MAE & HSIC-HSIC & MEE-MEE \\
\hline
\\[-6pt]
BKT &  \underline{0.16 $\pm$ 0.02} & \underline{0.16 $\pm$ 0.01} & 0.26 $\pm$ 0.07 & \underline{0.16 $\pm$ 0.014} \\
NT1 & 0.47 $\pm$ 0.01 & 0.48 $\pm$ 0.01 & 0.48 $\pm$ 0.01 & \underline{0.45 $\pm$ 0.007} \\
NT2 &  0.59 $\pm$ 0.01 & \underline{0.57 $\pm$ 0.01} & 0.59 $\pm$ 0.01 & 0.59 $\pm$ 0.013 \\
NT3 &  0.45 $\pm$ 0.01 & 0.45 $\pm$ 0.01 & 0.45 $\pm$ 0.01 & \underline{0.44 $\pm$ 0.005} \\
PMT &  \underline{0.44 $\pm$ 0.04} & 0.45 $\pm$ 0.05 & 0.52 $\pm$ 0.04 & 0.47 $\pm$ 0.021 \\
\hline
\\[-6pt]
\end{tabular}
\end{center}
\end{table}

\subsubsection{Comparison with State-of-the-art Approaches} % checked
Additionally, we compare our approach with two other state-of-the-art transfer learning algorithms tailored for regression: TrAdaBoost.R2 (TRB) \cite{pardoe_2010_ICML} and WANN \cite{mathelin_2021_IEEE_ICTAI}.
TRB is an ensemble learning method while WANN is based on adversarial learning.
We describe both approaches and their hyperparameters in detail in the supplementary material.

In the results shown in Table \ref{tab:sota_tl_results}, our fine-tuning approach using MEE is able to outperform the WANN in all tasks, and it performs significantly better than TRB in 4 out of 5 tasks.
This shows that fine-tuning is still a competitive method for time-series transfer learning.

It is important to emphasize that WANN and TRB were originally conceived for tabular data, so, besides our efforts to tune all hyperparameters to the time-series tasks, they might still be improved by more thorough adaptation.
It could be an interesting future work direction to adapt specialized transfer learning regression algorithms such as WANN and TRB to time-series tasks and to use MEE instead of MSE.

\begin{table}
\caption{Target squared error of fine-tuning using MEE and other SOTA transfer learning regression methods.}
\label{tab:sota_tl_results}
\begin{center}
\begin{tabular}{lllll}
\hline
\rule{0pt}{12pt}
\textbf{Data} & MSE-MSE & TRB & WANN & MEE-MEE \\
\hline
\\[-6pt]
BKT & \underline{0.16 $\pm$ 0.016} & 1.2 $\pm$ 1.2 & 0.48 $\pm$ 0.043 & \underline{0.16 $\pm$ 0.014} \\
NT1 & 0.47 $\pm$ 0.009 & 0.53 $\pm$ 0.011 & 0.94 $\pm$ 0.03 & \underline{0.45 $\pm$ 0.007} \\
NT2 & \underline{0.59 $\pm$ 0.014} & 0.66 $\pm$ 0.007 & 0.68 $\pm$ 0.095 & \underline{0.59 $\pm$ 0.013} \\
NT3 & 0.45 $\pm$ 0.008 & 0.52 $\pm$ 0.004 & 0.77 $\pm$ 0.018 & \underline{0.44 $\pm$ 0.005} \\
PMT & 0.44 $\pm$ 0.044 & \underline{0.38 $\pm$ 0.018} & 1.2 $\pm$ 0.49 & 0.47 $\pm$ 0.021 \\
\hline
\\[-6pt]
\end{tabular}
\end{center}
\end{table}

\section{Conclusion}
In this paper, we revisit the robustness of the MEE loss function to show that, besides its resilience to non-Gaussian response variable noise, it is also robust to covariate shift, a common challenge in many machine learning applications.
We draw awareness to the fact that MEE is better for training machine learning models for regression tasks than the commonly used MSE, and can even outperform state-of-the-art loss functions such as the HSIC.
We validate our hypothesis empirically on synthetic data representing different degrees of covariate shift.
Additionally, we show that MEE in combination with fine-tuning can outperform other loss functions and even other state-of-the-art transfer learning regression methods in real-world time-series tasks.

There are many future work possibilities based on our results about the robustness of the MEE loss.
Existing transfer learning algorithms such as TRB and WANN can be adapted to use MEE instead of MSE.
Our results also suggest that the out-of-distribution transfer learning generalization might also be improved by using MEE, which might lead to new interesting theoretical studies.

\ack This work has been conducted as part of the Just in Time Maintenance project funded by the European Fund for Regional Development.

\bibliography{main}

\appendix

\section{Robustness of MEE to Non-Gaussian Noises}

We just present here two insights on the robustness of $\min H(e)$ over the mean square error (MSE) criterion $\min E(e^2)$ against non-Gaussian noises, in which $E$ denotes the expectation. Interested readers can refer to~\cite{chen_AAS2009,chen_SP2010,chen_TNNLS18,hu_JMLR13} for more thorough analysis on the advantage of $\min H(e)$.

First,~\cite[Theorem~3]{chen_AAS2009} suggests that $\min E(e^2)$ is equivalent to minimizing the error entropy plus a Kullback–Leibler (KL) divergence. Specifically, we have:
\begin{equation}\label{eq:mse_mee}
    \min E(e^2)\Leftrightarrow \min H(e) + D_{\text{KL}}(p(e)\|\varphi(e)),
\end{equation}
in which $p(e)$ is the probability of error, $\varphi(e)$ denotes a zero-mean Gaussian distribution. As the KL-divergence is always nonnegative, minimizing the MSE is equivalent to minimizing an upper bound of the error entropy. Eq.~(\ref{eq:mse_mee}) also explains (partially) why in linear and Gaussian cases, the MSE and MEE are equivalent~\cite{kalata1979linear}. Nevertheless, in case the error or noise follows a highly non-Gaussian distribution (especially when the signal-to-noise (SNR) value decreases), the MSE solution deviates from the MEE result, but the latter takes full advantage of high-order information of the error~\cite{chen_TNNLS18}.

On the other hand, given the mean-square error $E(e^2)$, the error entropy satisfies~\cite{cover1999elements}:
\begin{equation}
    H(e)\le\max_{E(\zeta^2)=E(e^2)} H(\zeta)=\frac{1}{2}+\frac{1}{2}\log{2\pi}+\frac{1}{2}\log{(E(e^2))},
\end{equation}
where $\zeta$ denotes a random variable whose second moment equals to $E(e^2)$. This implies that the MSE criterion can be recognized as a robust MEE criterion in the minimax sense, because:
\begin{equation}
\label{eq_mee_upper}
\begin{split}
f_{\text{MSE}}^* & = \underset{{f\in F}}{\arg \min} E(e^2) \\
& = \underset{{f\in F}}{\arg \min} \frac{1}{2}+\frac{1}{2}\log{2\pi}+\log{(E(e^2))} \\
& = \underset{{f\in F}}{\arg \min} \max_{E(\zeta^2)=E(e^2)}H(\zeta),
\end{split}
\end{equation}
where $f_{\text{MSE}}^*$ denotes the solution with MSE criterion, $\mathcal{F}$ stands for the collection of all Borel measurable functions. Eq.~(\ref{eq_mee_upper}) suggests that minimizing the MSE is equivalent to minimizing an upper bound of the error entropy.

\section{Organization of the Related Work}
In Table \ref{tab:rel_work} we compare the domain generalization (DG) topic and more recent literature on new loss functions such as MEE and HSIC \cite{shujian_AAAI21,greenfield_icml2020,subbaswamy_aistats19,rothenhausler_jrssb21}.
The recent works on MEE, HSIC, and others, which inspire our work, are similar to DG in the sense that they assume that no data from the domain they aim to generalize to is available during training.
However, DG assumes that data from multiple additional domains are available during training, while MEE/HSIC papers assume training data to come from only one single domain.
% Domain generalization differs from the newer literature since it assumes multiple domains during training and the goal is to generalize to a new unseen domain.
% Furthermore, it does not have access to this new domain during training, while 

\begin{table*}[ht]
\caption{Comparison of related topics according to which data is accessible during training and testing, and the number of data domains.}
\label{tab:rel_work}
\begin{center}
\begin{tabular}{llll}
\hline
\rule{0pt}{12pt}
   &     Training data & Test data & Test access 
\\
\hline
\\[-6pt]
 Transfer Learning     & $\{\xvar_{iS}, y_{iS}\}_{i=1}^{N_S}, \{\xvar_{iT}, y_{iT}\}_{i=1}^{N_T}$   & $\{\xvar_{iT}, y_{iT}\}_{i=1}^{N_T}$  & yes \\
 Domain Adaptation     & $\{\xvar_{iS}, y_{iS}\}_{i=1}^{N_S}, \{\xvar_{iT}\}_{i=1}^{N_T}$   & $\{\xvar_{iT}, y_{iT}\}_{i=1}^{N_T}$  & yes \\
 Domain Generalization & $\{\xvar_{i1}, y_{i1}\}_{i=1}^{N_1}, \{\xvar_{i2}, y_{i2}\}_{i=1}^{N_2}, ... \{\xvar_{in}, y_{in}\}_{i=1}^{N_n}$    & $\{\xvar_{in+1}, y_{in+1}\}_{i=1}^{N_{n+1}}$   & no \\ 
 MEE/HSIC              & $\{\xvar_{i1}, y_{i1}\}_{i=1}^{N_1}$   & $\{\xvar_{i2}, y_{i2}\}_{i=1}^{N_2}$  & no \\ 
     % \hline
% MEE  & $\sigma_Y$ &  1 & 0.1 &  1 &  0.3  &  0.8 &  1 & 0.5 &  0.3  \\

\hline
\\[-6pt]
% \multicolumn{4}{c}{$\mathcal{D}_i=\{\xvar_{j}, y_{j}\}_{j=1}^{N_i}$}
\end{tabular}
\end{center}
\end{table*}

\section{Implementation Details of the Synthetic Data Experiment}
We approximate the parameters $\theta$ through linear regression by minimizing the MEE and the other losses by using the gradient descent optimization algorithm Adam \cite{kingma_adam}.
We run the optimization for 500 epochs, batch size of 128 samples, and learning rate of $10^{-4}$.
The source and target dataset sizes are both 1000 samples, and we repeat the experiment 100 times for each version of $p_T(x)$ and $\epsilon$.
In order to also compare the robustness of the losses to different noise distributions, we repeat the experiment three times, one where $\epsilon$ follows a shifted exponential distribution with $\lambda=1$, another where it follows a zero-centered Laplace distribution with scale $b=1$, and, finally, using mixed Gaussian noise: $\epsilon \sim 0.95\mathcal{N}(0, 0.01) + 0.05 \mathcal{N}(0, 100)$.
The HSIC uses RBF kernels for the covariates and the residuals and the kernel size is 1 for both, as implemented in a similar experiment from the authors \cite{greenfield_icml2020}.
We use the same kernel size for MEE.

\section{Hyperparameter Choice for the TCN}
We parameterize the TCN with one residual block of 128 filters, kernel size of 3, and a dropout rate of $0.1$. 
The residual block is composed of one skip connection and two diluted causal convolutions, the first one with dilation of 1 and the second with dilation of 2.
% This final architecture was selected through grid search with scope defined in Table \ref{todo} in the Appendix.
The last time-step outputted by the last convolutional layer is inputted into a fully connected layer to give the final output.
The TCN is trained for 200 epochs using a learning rate of $10^{-4}$ and batch size of 64.
We use 10\% of the training data for validation and we store the best weights during training according to the validation error to mitigate overfitting.
The hyperparameters were selected through Bayesian optimization and were similar for all the datasets.

\section{Dataset Details}
The Nasa Turbofan data concerns predicting the remaining useful life of turbofan engines by using sensor information extracted from them.
It is split into 4 datasets, each one containing sensor measurements from operational cycles of 100 engines running in different environmental conditions and fault modes.
We select the source (NTS) to be the dataset where all engines operate in the same condition and fault mode and we take the data from 80 engines to be the training set and the rest for validation.
From each of the three other datasets (NT1, NT2, and NT3) we separate 10 engines for the target training set and we use the rest for testing.
The engines in each target dataset can operate in different combinations of environmental conditions and fault modes, providing covariate shift with respect to the source dataset NTS:
\begin{itemize}
    \item NT1 has the same fault mode as NTS, but has six different environmental conditions;
    \item NT2 has the same environment condition as NTS, but has one different fault mode;
    \item NT3 has different fault modes and environmental conditions than NTS.
\end{itemize}
For each engine, we extract time windows of 30 cycles and we use the 14 most informative sensors as also done in other works \cite{nasa_dataset_win_size}. 
The prediction label is the number of operational cycles left until the engine fails at the end of the time window.

The Beijing air quality dataset contains hourly meteorological measurements from 12 stations in the period between 2013 and 2017, and the goal is to predict the concentration of PM10 in the air.
The input is composed of windows of 24 hours and 9 measurements including wind speed, temperature, precipitation, and concentrations of CO, O3, NO2, and SO2, and the label is the PM10 concentration at the end of the window period.
The source dataset (PMS) is extracted from the first year of measurements (2013) while the target data (PMT) is from the last year (2016 to 2017), so we can expect a distribution change in the observed concentration of gases between 2013 and 2016.
The training target data only includes the early months in the period selected by PMT.

The bike sharing dataset collects one year of hourly meteorological data combined with the number of bikes rented by a bike sharing company.
We build the time-series dataset from it by sampling overlapping time windows of 24 hours and labeling them with the renting amount at the end.
Only the meteorological information is used as input for the models.
The source dataset is sampled from the days during fall, winter, and spring, while the target data is from summer, and the training target dataset contains information only from the early days of summer.
Here the covariate shift is represented in the form of meteorological changes between seasons.

Besides the aforementioned discrepancies between source and target datasets, another potential factor of covariate shift is the difference in size between the target training sets and the target test sets.
The size of the target training sets is limited compared to the target test set, so it is likely that only part of the input distribution will be represented in the training set.

\begin{table*}[ht]
\caption{Comparing noise robustness of MEE with Huber and Jonathan T. losses.}
\label{tab:noise_comparison}
\begin{center}
\begin{tabular}{llllll}
\hline
\rule{0pt}{12pt}
    \textbf{Noise} &  MAE & HSIC & Huber & Jonathan T. & MEE \\

\hline
\\[-6pt]
Laplace & 3.46 $\pm$ 1.66 & 3.81 $\pm$ 2.25 & 3.93 $\pm$ 2.46  & 5.73 $\pm$ 3.56 & 3.58 $\pm$ 1.81 \\
Mix. Gaussian & 32.66 $\pm$ 7.37 & 30.49 $\pm$ 5.42 & 32.88 $\pm$ 7.85 & 29.71 $\pm$ 4.89 & 29.50 $\pm$ 4.57 \\
\hline
\\[-6pt]
\end{tabular}
\end{center}
\end{table*}

\section{Description and Hyperparameter Choice of WANN and TRB}
TRB is an ensemble transfer learning method that combines both source and target training samples and weighs each sample during training according to the error of each learner on that sample.
Similar to any boosting algorithm, it trains multiple instances of a base learner and weights the training data according to the error obtained from the learners.
However, for the source samples, the weighting is reversed, so more difficult source samples get lower weights. 
The idea is that difficult source samples are shifted further away from the target data distribution, so minimizing their importance during training ensures that the ensemble will be more robust to the shifts in the transfer learning task.
In our experiments, we use the two-step variation of TRB in which an initial boosting step is performed to compute only the target weights, and then both source and target weights are tuned simultaneously in a subsequent step.
This is done to prevent the problem of vanishing source weights spotted by the authors.

WANN is a neural network trained to minimize an upper bound of the target generalization error based on the $\mathcal{Y}$-discrepancy \cite{y-discrepancy_ICALT2012} between the source and the target distributions.
Since the $\mathcal{Y}$-discrepancy is computed as the maximum difference between source and target error over all possible hypothesis $h$, the loss minimized by WANN becomes an adversarial function:
\begin{equation*}
    \begin{split}
    \min_{h, q} \max_{h'} 
    \sum_{x_S, y_S}q(x_S)\mathcal{L}(h(x_S), y_S) + & \sum_{x_T, y_T} \mathcal{L}(h'(x_T), y_T) \\
                                                  - & \sum_{x_S, y_S} q(x_S) \mathcal{L}(h'(x_S), y_S) 
    \end{split}
\end{equation*}
% \begin{equation*}
%     \min_{h, q} \max_{h'} 
%     \sum_{x_S, y_S}q(x_S)\mathcal{L}(h(x_S), y_S) + \sum_{x_T, y_T} \mathcal{L}(h'(x_T), y_T) - \sum_{x_S, y_S} q(x_S) \mathcal{L}(h'(x_S), y_S) 
% \end{equation*}
where $h$ and $h'$ are neural network predictors and $q$ is a neural network that outputs a weight for each source sample.
The core idea behind it is that by minimizing the $\mathcal{Y}$-discrepancy, the weighting network $q$ will help to correct the shift between source and target domains, so the final learner $h$ will generalize better to the target domain.
We use the implementations of WANN and TRB provided by the Adapt Python package \cite{de2021adapt_package}.

We run WANN using a TCN for both the predictor and the weighting network, and we use the same hyperparameters selected for the classic transfer learning approaches.
This architecture should be appropriate since it shows overall better performance in the source datasets, as recommended by the authors \cite{mathelin_2021_IEEE_ICTAI}.
We select the weight clipping parameter as 85, a dropout rate of 0.3, a learning rate and L2 regularization of $10^{-5}$.
The WANN is trained for 300 epochs and uses early stopping based on the error computed on 10\% of the target training data which is left out during training.
For TRB, we select the base learner to be a multi-layer perceptron (MLP) with three hidden layers of 1024 units, since training using the TCN architecture was impracticable due to the high memory demand of having an ensemble of TCNs.
The MLP is parameterized with a dropout rate of 15\%, a learning rate of $10^{-3}$, and L2 regularization of $10^{-4}$ and also runs with early-stopping based on the validation data.
To make the data compatible with the MLP learners, we extract four time-independent features from each channel of the time-series inputs: mean, standard deviation, min, and max values.
The TRB algorithm is parameterized with 9 learners for the first stage and 12 learners for the second stage and 5 cross-validation iterations.
The hyperparameters are selected through Bayesian optimization on each transfer learning task and the results were similar for all tasks.

\section{Additional Experiments}
Hereby we present the results of the additional experiments comparing our approach with other robust loss functions and transfer learning methods.

\subsection{Comparison with other noise-robust losses}
In addition to the experiments presented in the main paper, here we compare the MEE with two other noise-robust loss functions: the Huber loss and the Jonathan T. loss \cite{Barron_2019_CVPR}.
One theoretical disadvantage of both losses with respect to MEE is that they focus on robustness to noises or outliers, rather covariate shift.
Another advantage of MEE is that it does not have any additional hyperparameter to tune, while the Huber and Jonathan have respectively one and two hyperparameters that require manual tuning.
That is why we prefer the MEE over Huber and Jonathan T for our transfer learning method.

In Table \ref{tab:noise_comparison} we show the comparison result using the same covariate shift simulation experiment data from the main paper using Laplace and mixed gaussian noises where the shift is 2. We parameterize the Huber loss with delta of 4 and the Jonathan T. with scale of -4 and delta of 0.1. The values were selected through grid search on the unshifted training data.
The result shows that MEE outperforms both losses in both experiments.

\subsection{Comparison with TCA}
Here we compare our approach with a more traditional transfer learning method, the transfer component analysis (TCA) \cite{pan_2011_IEEE_trans_NN} using the PMT and BKT datasets. 
The TCA focuses on tackling the distribution shift problem directly by casting both source and target domains to a subspace where their distributions are closer.
In order to apply the TCA to the time-series data, we first process it to the same non-temporal features used for TRB since it is a linear method and cannot properly capture the time dimension.

Table \ref{tab:tl_comparison} shows the result of the comparison between TRB, WANN, TCA, and our fine-tuning approach combined with MEE.
In comparison with other more recent transfer learning methods such as TRB and WANN, the TCA shows mixed results but never outperforms both of them simultaneously.
Finally, when compared with our approach the TCA shows significantly lower performance, proving that our method is better at handling covariate shift in complex data such as time series.

\begin{table}
\caption{Comparing our approach with TCA on real-life transfer learning tasks.}
\label{tab:tl_comparison}
\begin{center}
\begin{tabular}{lllll}
\hline
\rule{0pt}{12pt}
    \textbf{Noise} &  TRB & WANN & TCA & MEE-MEE \\

\hline
\\[-6pt]
PMT & 0.38 $\pm$ 0.018 & 1.2 $\pm$ 0.49 & 0.59 $\pm$ 0.05  & 0.47 $\pm$ 0.021  \\
BKT & 1.2 $\pm$ 1.2  & 0.48 $\pm$ 0.043 & 0.72 $\pm$ 0.02 & 0.16 $\pm$ 0.014 \\
\hline
\\[-6pt]
\end{tabular}
\end{center}
\end{table}

\end{document}